\documentclass{article}

\usepackage{PRIMEarxiv}

\usepackage[T1]{fontenc}
\usepackage{amsmath, amssymb}
\usepackage{array}
\usepackage{multirow}
\usepackage{booktabs}
\usepackage{placeins}
\usepackage{graphicx}
\usepackage{microtype}
\usepackage{caption}
\usepackage{csquotes}
\newcommand*{\nowbiblatexbackend}{biber}
\usepackage{now-sip-biblatex}
\AtBeginDocument{\renewbibmacro*{journal+issuetitle}{\usebibmacro{journal}\setunit*{\addspace}\iffieldundef{series}{}{\newunit\printfield{series}\setunit{\addspace}}\usebibmacro{volume+number+eid}\setunit{\addspace}\usebibmacro{issue+date}\setunit{\bibpagespunct}\printfield{pages}\newunit}}
\appto{\bibsetup}{\emergencystretch=3em}
\DeclareUnicodeCharacter{1EF3}{\`y}
\DeclareUnicodeCharacter{0301}{\'{e}}

\usepackage[colorlinks=true, allcolors=blue]{hyperref}
\hypersetup{pdftitle={What Does Each Part of Speech Need? Contexts, Subspaces and Dimensions in Static Word Embeddings}, pdfauthor={Jintang Xue, Chengwei Wei, Yun-Cheng Wang, C.-C. Jay Kuo}, pdfkeywords={static word embeddings, part of speech, per-POS PCA, dimensionality reduction, dimension allocation}}

\title{What Does Each Part of Speech Need?\\ Contexts, Subspaces and Dimensions\\ in Static Word Embeddings}

\usepackage{authblk}
\author[1]{Jintang Xue}
\author[1]{Chengwei Wei}
\author[1]{Yun-Cheng Wang}
\author[1]{C.-C.~Jay~Kuo}
\affil[1]{University of Southern California, Los Angeles, California, USA}

\begin{document}
\maketitle

\begin{abstract}
Static word embeddings give a word one vector, ignoring its different functions as noun, verb or adjective. Training on part-of-speech-tagged text gives each function its own vector, at the cost of a larger table. Using nine controlled corpora, we ask which context words each part of speech needs, whether it needs its own subspace, and whether the number of dimensions a word keeps should follow its part of speech or its frequency. The tags raise verb similarity on SimVerb-3500 at every dimension from 50 to 300 but help nouns much less. Added to verb-only text, adverbs help verbs more than as many random words; adjectives help nouns only slightly. At a fixed dimension, per-part-of-speech PCA bases preserve verb and adjective similarity better than a global basis or frequency-band bases. Nouns need more dimensions than verbs only among frequent words. We therefore allocate dimensions by frequency with these bases: at a third of the original size, only rare words lose dimensions, SimVerb-3500 and SimLex-999 nouns match the uncompressed table, and verb-to-noun retrieval stays within 0.02. Part of speech decides what a word's vector should be; frequency decides how much of it to keep.
\end{abstract}

\FloatBarrier
\section{Introduction}\label{sec:intro}

Language models and the chatbots built on them now dominate natural language processing \cite{wei2023overview,xue2023bias}, but static word embeddings, such as word2vec \cite{mikolov2013efficient,mikolov2013distributed}, GloVe \cite{pennington2014glove}, and fastText \cite{bojanowski2017enriching}, remain in use where models must be small and fast \cite{kuo2023green,wang2023overview}. A static embedding is looked up in a table rather than computed by a network over the sentence, and that table is the main size bottleneck of many NLP models \cite{acharya2019online}; text classifiers built on such tables can be compressed to fit in a limited amount of memory \cite{joulin2016fasttext}, and static tables are now also obtained from contextual models \cite{bommasani2020interpreting,gupta2021obtaining,tulkens2024model2vec}.

A static table, however, gives each word one vector for all its meanings \cite{camacho2018word}: the verb \textit{run} and the noun \textit{run} share one vector. Training on text tagged with part-of-speech (POS) tags, as sense2vec \cite{trask2015sense2vec} does, gives the two separate vectors, each learned only from the contexts of its own part of speech. Separating them helps: POS-tagged vectors reduce the unlabeled attachment errors of a neural dependency parser by more than 8 percent on average across six languages \cite{trask2015sense2vec}, and in our experiments the tags raise similarity on SimVerb-3500 at every dimension from 50 to 300 (Section~\ref{sec:tags}). But the tags also make the table larger: a 300-dimensional tagged table trained on English Wikipedia has 2.06 million rows, one vector for each word and POS tag, and takes 2,468 MB in float32. Methods that reduce the size of such tables usually give frequent words more dimensions than rare ones \cite{chen2018groupreduce,baevski2018adaptive,ginart2021mixed} and ignore part of speech, although verbs, nouns and adjectives behave differently, and verbs in particular are a known weak point of static embeddings \cite{gerz2016simverb}.

We ask what each part of speech needs, in three steps. During training, we ask whether POS tags help a POS class and which nearby words it needs as context, for example whether verbs need their adverbs. After training, we ask whether each POS class occupies directions of its own in the vector space and how many dimensions it needs. Finally, we ask how a budget of dimensions should be spent across words, by part of speech or by frequency.

\begin{figure}[!t]
\centering
\includegraphics[width=\linewidth]{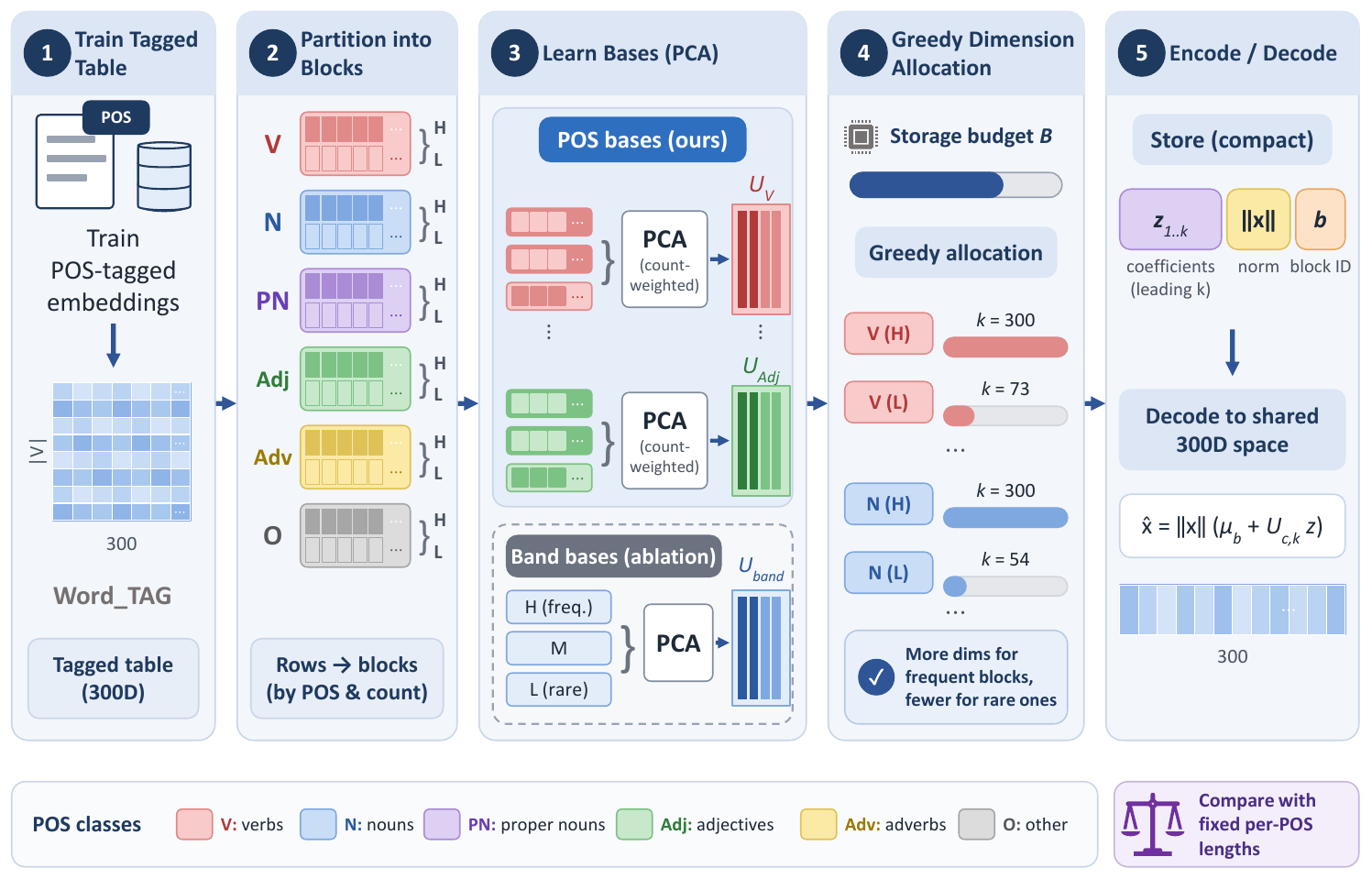}
\caption{The pipeline that Section~\ref{sec:method} tests, in five steps. (1) One tagged table is trained at 300 dimensions. (2) Its rows are grouped into blocks by part of speech and training count, with the bands cut at $10^2$, $10^3$, $10^4$ and $10^5$ (H and L: the most and least frequent blocks of a class), and each block gets a count-weighted mean. (3) The blocks of a part of speech share its count-weighted PCA basis $U_c$ (the POS bases, Section~\ref{sec:weighting}); as an ablation, each frequency band is one block with its own basis (the band bases). (4) A budget of dimensions is spent across the blocks by a greedy rule that favors frequent blocks (Section~\ref{sec:method-def}); the numbers of directions shown are those of the allocation at a mean of 100 per row for the most and least frequent verb and noun blocks. (5) Each word stores its first $k$ coefficients $z$, its norm $\|x\|$ and its block $b$, and is decoded into the shared 300-dimensional space from the block mean $\mu_b$ and the first $k$ columns of its class basis. Section~\ref{sec:method-results} compares this with one length per part of speech.}\label{fig:method}
\end{figure}

We answer with controlled corpora (Section~\ref{sec:protocol}): nine corpora cut from one POS-tagged Wikipedia dump, from the full text with and without POS tags to corpora that keep only some parts of speech, such as verbs alone or verbs with their adverbs. Each added POS class has a baseline that adds as many random words from the same lines. Every model is trained with three seeds and two context windows. Figure~\ref{fig:method} shows the pipeline of the last step: the rows of the tagged table are grouped into blocks by part of speech and training count, and a budget of dimensions is spent across the blocks.

We find that part of speech decides what a word's vector should be, through the contexts it is trained on and, where rows are cut to one length, the subspace that describes it, while how many dimensions a word should keep is set by its frequency. Our contributions are as follows.

\begin{itemize}
\item \textbf{Controlled corpora.} Nine corpora cut from one tagged Wikipedia dump, with and without POS tags or restricted to some parts of speech, isolate one factor at a time (Section~\ref{sec:protocol}).

\item \textbf{What each POS class needs during training.} POS tags consistently help verbs and help nouns much less, and the vectors must stay separate: a word represented by the average of its vectors falls below the untagged table on MEN with all pairs. Verbs gain more from their adverbs than from random words, whereas adjectives give nouns only a small gain. We therefore train a single table on the fully tagged corpus and keep one vector per word and part of speech (Sections~\ref{sec:tags} and \ref{sec:context}).

\item \textbf{What each POS class needs in the trained space.} We split the tagged table into six POS classes: verbs, nouns, proper nouns, adjectives, adverbs and other words. At the same reduced dimension, a separate PCA for each POS class preserves verb and adjective similarity better than a single PCA and keeps verbs and nouns closer to semantically related words. Among frequent words, nouns need more dimensions than verbs, and verbs more than other words (Section~\ref{sec:geometry}).

\item \textbf{Spending a budget of dimensions.} How many dimensions a word keeps should follow its frequency: with the POS bases and a greedy allocation by training count, from a third of the table's storage on, every word seen at least 100 times keeps all 300 dimensions, SimVerb-3500 and SimLex-999 nouns stay at the full table's level and verb-to-noun retrieval stays within 0.02. Nouns need more dimensions than verbs only among frequent words (Section~\ref{sec:geometry}), so no part of speech needs a length of its own. Against one basis per frequency band, the POS bases keep up to 0.04 more of the verb-to-noun retrieval and lose only on the rare words (Section~\ref{sec:method}).
\end{itemize}

The rest of this paper is organized as follows. Section~\ref{sec:related} reviews related work, and Section~\ref{sec:protocol} describes the corpora and evaluation. Sections~\ref{sec:tags} to \ref{sec:method} report the results, and Section~\ref{sec:conclusion} concludes.

\section{Related Work}\label{sec:related}

\textbf{POS tags and contexts.}
Training word2vec \cite{mikolov2013efficient,mikolov2013distributed} on tokens that join each word with its POS tag, as sense2vec \cite{trask2015sense2vec} does, gives one row per word and POS tag. Levy and Goldberg \cite{levy2014dependency} used dependency contexts, which favor functional over topical similarity. Others chose contexts per part of speech, through dependency configurations \cite{vulic2017automatic} or symmetric patterns for verbs and adjectives \cite{schwartz2016symmetric}, or made them position-aware \cite{ling2015two}. Our corpora instead isolate the POS tags and each added POS class (Section~\ref{sec:corpora}).

\textbf{Dimension reduction.}
Earlier methods pick one number of dimensions for the whole table \cite{yin2018dimensionality}, remove the mean and top principal components \cite{mu2017all}, combine that post-processing with PCA \cite{raunak2019effective}, or select dimensions with weak supervision \cite{xue2024word}. Kutuzov et al. \cite{kutuzov2016redefining} find POS information spread over many components, and Musil \cite{musil2019examining} relates principal components to part of speech, but neither asks how many dimensions each POS class needs or whether that number can be used to reduce the table. We use per-POS PCA, centered on the count-weighted mean of each POS class. Our earlier method \cite{xue2024word} selects dimensions of one untagged table with the supervised feature-selection tests of \cite{yang2022supervised} and treats all words alike; here the table is tagged, and we ask what each part of speech needs.

\textbf{Smaller tables and sense vectors.}
Other ways to make a table small keep its dimensions and change how a row is stored, as compositional codes \cite{shu2017compressing} and binarization \cite{tissier2019near} do, or give frequent words more dimensions than rare ones \cite{chen2018groupreduce,baevski2018adaptive,ginart2021mixed}, which Section~\ref{sec:method} builds on. Sense embeddings give a word one vector per sense induced from its contexts \cite{huang2012improving,neelakantan2014efficient}; a POS tag is a coarser split that needs no induction, one vector per part of speech.

\textbf{Dimensions by group.}
GroupReduce \cite{chen2018groupreduce} gives each frequency group its own low-rank approximation, with higher rank for frequent words. Mixed-dimension embeddings \cite{ginart2021mixed} and adaptive input representations \cite{baevski2018adaptive} also favor frequent rows and decode to one size, and Matryoshka representation learning \cite{kusupati2022matryoshka} makes every prefix of a vector usable for every word. The closest classical method, local PCA \cite{kambhatla1997dimension}, fits one PCA per partition of the data, with lower error than global PCA at the same dimension; adaptive transform coding \cite{dony1995optimally} is similar. Their partitions are learned, whereas ours is given by the POS tag.

Per-group PCA, greedy allocation and decoding to a shared space are standard, but dimension selection has mostly treated the table as one population or split it by frequency. What we add is part of speech as the partition, compared with frequency bands at equal dimension (Section~\ref{sec:subspaces}) and tested as a rule for how many dimensions each class keeps (Section~\ref{sec:method}).

\FloatBarrier
\section{Experimental Setup}\label{sec:protocol}

\subsection{Corpora}\label{sec:corpora}

\begin{figure}[!t]
\centering
\includegraphics[width=\linewidth]{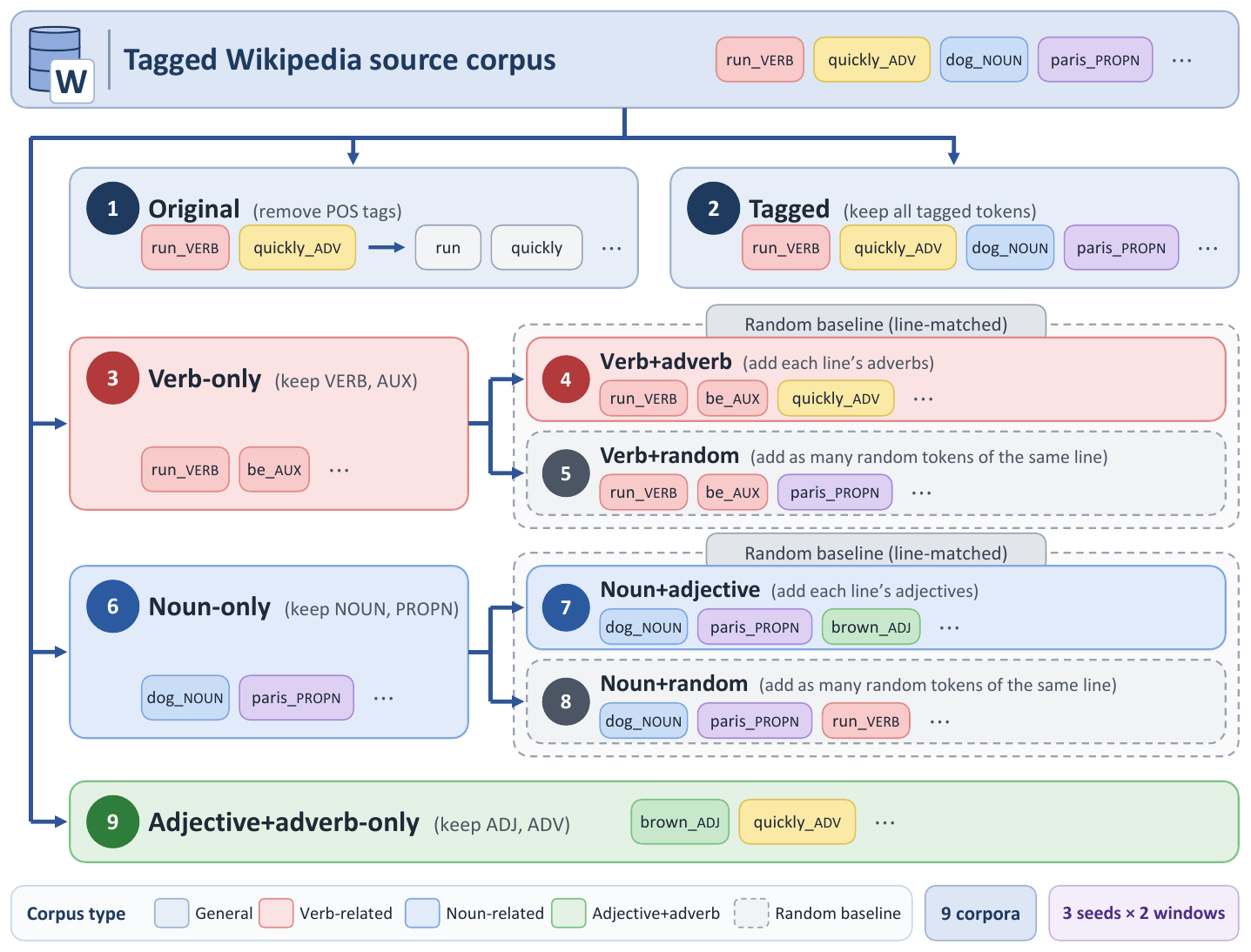}
\caption{How the nine corpora are derived, numbered in the order of Table~\ref{tab:corpora}. All nine are cut from one tagged Wikipedia corpus, and corpora joined by an arrow differ in one factor; the boxes under each name are example tokens, lowercased lemmas with their POS tags. Each random baseline (dashed) keeps the lines of the corpus it is matched to and, in every line, draws as many random tokens as the added POS class has there. Every corpus is trained with three seeds and two windows.}\label{fig:chain}
\end{figure}

We start from the English Wikipedia dump of November 1, 2025, tagged by spaCy 3.8.14 \cite{honnibal2020spacy} with the \texttt{en\_core\_web\_sm} model and universal POS tags. Every token is lowercased, lemmatized and written as \texttt{word\_TAG}, and a fixed list of common prepositions is removed as stop words. The lines of this file are chunks of running text about 30 tokens long, not sentences, and no training window crosses a line. Each corpus (Table~\ref{tab:corpora}, Figure~\ref{fig:chain}) deletes tokens inside the lines of this file and drops lines left empty. Noun-only keeps PROPN tokens as well as NOUN tokens: proper nouns fill the same positions as nouns and supply nearly half of its tokens, and Section~\ref{sec:geometry} treats their rows as a class of their own. Verb+adverb and noun+adjective add each line's adverbs or adjectives back to verb-only and noun-only, and their random baselines add as many random tokens of the same line, kept in order. The original and tagged corpora show what POS tags add (Section~\ref{sec:tags}). The filtered corpora, each trained separately, show which context words a POS class needs and whether it does better with its own table (Section~\ref{sec:context}). We reduce only the tagged table (Sections~\ref{sec:geometry} and \ref{sec:method}).

\begin{table}[tp]
\begin{center}
\caption{The nine corpora. Kept: the POS tags a corpus retains; +random: as many random tokens of the same line in place of the adverbs or adjectives. Vocabulary: distinct tokens seen at least five times, each a row of the trained table; the tagged corpus has more because a word gets one row per POS tag (run\_VERB, run\_NOUN). Each random baseline has slightly fewer tokens than its counterpart (1,197 for verb+random, 377,674 for noun+random), because a line with too few other tokens gives fewer random tokens, and three fewer lines, the source lines made only of adverbs or adjectives.\label{tab:corpora}}
\small
\setlength{\tabcolsep}{3pt}
\renewcommand{\arraystretch}{1.15}
\begin{tabular*}{\linewidth}{@{\extracolsep{\fill}}llrrrr@{}}
\toprule
\textbf{Corpus} & \textbf{Kept} & \textbf{Lines} & \textbf{Tokens} & \textbf{Tokens/line} & \textbf{Vocabulary} \\
\midrule
original & all, no POS tags & 81,972,758 & 2,459,902,070 & 30.01 & 1,765,035 \\
tagged & all, tagged & 81,972,758 & 2,459,902,070 & 30.01 & 2,056,807 \\
\midrule
verb-only & VERB, AUX & 81,492,773 & 401,914,303 & 4.93 & 52,459 \\
verb+adverb & +ADV & 81,570,119 & 475,397,674 & 5.83 & 68,625 \\
verb+random & +random & 81,570,116 & 475,396,477 & 5.83 & 306,155 \\
\midrule
noun-only & NOUN, PROPN & 81,970,573 & 1,077,465,222 & 13.14 & 1,821,083 \\
noun+adjective & +ADJ & 81,971,070 & 1,274,151,854 & 15.54 & 1,935,638 \\
noun+random & +random & 81,971,067 & 1,273,774,180 & 15.54 & 1,862,214 \\
\midrule
adjective+adverb-only & ADJ, ADV & 77,040,811 & 270,170,003 & 3.51 & 130,721 \\
\bottomrule
\end{tabular*}
\end{center}
\end{table}

\subsection{Models and training}\label{sec:training}

All models are gensim \cite{vrehuuvrek2010software} skip-gram with negative sampling \cite{mikolov2013distributed}: 10 negative samples, minimum count 5, subsampling threshold $10^{-4}$, 5 epochs, windows $w{=}1$ and 5, and seeds 1, 2 and 3. Window, subsampling and similar settings explain much of the difference between methods \cite{levy2015improving}, so all models share this recipe and are compared within one window on the same pairs. A row of a trained table is the vector of one distinct token seen at least five times, so in the tagged table a word has one row for each POS tag it receives at least five times. Each corpus is trained at 50, 100, 200 and 300 dimensions. We give storage in MB of $10^{6}$ bytes.

\subsection{Benchmarks and evaluation}\label{sec:evaluation}

\textbf{Word similarity.} The main benchmarks are SimVerb-3500 \cite{gerz2016simverb} and SimLex-999 nouns \cite{hill2015simlex}. Verbs are also scored on SimLex-999 verbs and YP-130 \cite{yang2006word}, and nouns on MC-30 \cite{miller1991contextual}, RG-65 \cite{rubenstein1965contextual} and the noun subsets of MEN \cite{bruni2014multimodal}, WS-353 \cite{finkelstein2001placing}, MTurk-771 \cite{halawi2012large} and RW \cite{luong2013better}. These subsets keep only pairs whose words are tagged NOUN or PROPN in at least 90 percent of their occurrences. The adjective benchmark is SimLex-999 adjectives. Scores are Spearman correlations between cosine and gold ratings. A tagged model represents each word by its row for the benchmark's part of speech: VERB, NOUN or ADJ. A verb with no VERB row falls back to its AUX row, and a noun with no NOUN row to its PROPN row. The full MEN and WS-353 serve as mixed sets, where the model represents a word by the count-weighted average of all its rows. Word-similarity tasks have known limits \cite{faruqui2016problems,wang2019evaluating}, so the reduced-dimension experiments add the measures below.

\textbf{Other measures (reduced-dimension experiments).} Verb-to-noun retrieval uses 8,800 WordNet \cite{fellbaum1998wordnet} pairs of a verb and a noun derived from it, such as decide and decision, whose rows are both seen at least 20 times: the 5,646 distinct verb rows of these pairs each search all 129,677 noun rows by cosine, and we report the mean reciprocal rank (MRR) of the best-ranked derived noun. Rare-word neighbor retention (R1) is the share of the 10 nearest neighbors of 2,000 rows with counts of 20 to 100 that stay in the top 10 after reduction. Retrieval and neighbor retention use only rows seen at least 20 times. All scores are means over three seeds, and differences are reported without significance tests.

\FloatBarrier
\section{What Part-of-Speech Tags Contribute}\label{sec:tags}

We first test whether POS tags help each POS class at training time. Table~\ref{tab:tags} compares tagged and original tables of equal dimension.

\begin{table}[!t]
\begin{center}
\caption{Tagged minus original by dimension, in Spearman correlation averaged over 3 seeds, for windows $w{=}1$ and $w{=}5$. $n$ is the number of pairs. The original and tagged columns are the scores at 100 dimensions, with the higher of each pair in bold; $\Delta$ is tagged $-$ original at 50, 100, 200 and 300 dimensions. The tagged model uses the row of the benchmark's POS class. On MEN and WS-353 with all pairs the POS class is unknown, so it uses the count-weighted average of the word's vectors. $\Delta$ is computed before rounding.\label{tab:tags}}
\small
\setlength{\tabcolsep}{3pt}
\renewcommand{\arraystretch}{1.0}
\begin{tabular*}{\linewidth}{@{\extracolsep{\fill}}lrccccccc@{}}
\toprule
 & & & \multicolumn{2}{c}{100 dimensions} & \multicolumn{4}{c}{$\Delta$ at 50 to 300 dimensions} \\
\cmidrule(lr){4-5}\cmidrule(lr){6-9}
Benchmark & $n$ & $w$ & original & tagged & 50 & 100 & 200 & 300 \\
\midrule
\multicolumn{9}{l}{\textbf{Verbs}} \\
\multirow{2}{*}{SimVerb-3500} & \multirow{2}{*}{3,500} & 1 & 0.270 & \textbf{0.303} & $+0.026$ & $+0.033$ & $+0.033$ & $+0.029$ \\
 &  & 5 & 0.229 & \textbf{0.250} & $+0.016$ & $+0.021$ & $+0.026$ & $+0.024$ \\
\multirow{2}{*}{SimLex-999 verbs} & \multirow{2}{*}{222} & 1 & 0.221 & \textbf{0.235} & $+0.020$ & $+0.015$ & $+0.010$ & $+0.001$ \\
 &  & 5 & 0.140 & \textbf{0.146} & $-0.002$ & $+0.007$ & $+0.008$ & $+0.006$ \\
\multirow{2}{*}{YP-130} & \multirow{2}{*}{130} & 1 & 0.526 & \textbf{0.607} & $+0.106$ & $+0.080$ & $+0.073$ & $+0.064$ \\
 &  & 5 & 0.505 & \textbf{0.581} & $+0.099$ & $+0.076$ & $+0.070$ & $+0.057$ \\
\midrule
\multicolumn{9}{l}{\textbf{Nouns}} \\
\multirow{2}{*}{SimLex-999 nouns} & \multirow{2}{*}{666} & 1 & \textbf{0.427} & 0.423 & $0.000$ & $-0.004$ & $-0.007$ & $+0.002$ \\
 &  & 5 & \textbf{0.393} & 0.391 & $+0.002$ & $-0.002$ & $+0.005$ & $0.000$ \\
\multirow{2}{*}{MC-30} & \multirow{2}{*}{30} & 1 & 0.705 & \textbf{0.801} & $+0.122$ & $+0.096$ & $+0.009$ & $-0.005$ \\
 &  & 5 & 0.743 & \textbf{0.845} & $+0.083$ & $+0.102$ & $+0.039$ & $+0.036$ \\
\multirow{2}{*}{RG-65} & \multirow{2}{*}{65} & 1 & 0.798 & \textbf{0.844} & $+0.057$ & $+0.047$ & $+0.031$ & $+0.022$ \\
 &  & 5 & 0.792 & \textbf{0.838} & $+0.039$ & $+0.046$ & $+0.008$ & $+0.008$ \\
\multirow{2}{*}{MEN nouns} & \multirow{2}{*}{1,382} & 1 & 0.724 & \textbf{0.730} & $+0.011$ & $+0.006$ & $+0.006$ & $+0.005$ \\
 &  & 5 & 0.744 & \textbf{0.748} & $+0.009$ & $+0.004$ & $+0.001$ & $+0.001$ \\
\multirow{2}{*}{WS-353 nouns} & \multirow{2}{*}{272} & 1 & \textbf{0.677} & 0.668 & $-0.009$ & $-0.009$ & $-0.012$ & $-0.003$ \\
 &  & 5 & 0.713 & \textbf{0.714} & $-0.011$ & $+0.001$ & $+0.002$ & $-0.010$ \\
\multirow{2}{*}{MTurk-771 nouns} & \multirow{2}{*}{410} & 1 & 0.696 & \textbf{0.700} & $-0.001$ & $+0.004$ & $-0.010$ & $-0.018$ \\
 &  & 5 & \textbf{0.712} & 0.702 & $-0.005$ & $-0.009$ & $-0.018$ & $-0.023$ \\
\multirow{2}{*}{RW nouns} & \multirow{2}{*}{537} & 1 & 0.339 & \textbf{0.353} & $+0.002$ & $+0.014$ & $+0.018$ & $+0.018$ \\
 &  & 5 & 0.346 & \textbf{0.362} & $+0.006$ & $+0.016$ & $+0.022$ & $+0.030$ \\
\midrule
\multicolumn{9}{l}{\textbf{Adjectives}} \\
\multirow{2}{*}{SimLex-999 adjectives} & \multirow{2}{*}{111} & 1 & 0.455 & \textbf{0.514} & $+0.040$ & $+0.059$ & $+0.038$ & $+0.035$ \\
 &  & 5 & 0.391 & \textbf{0.443} & $+0.006$ & $+0.052$ & $+0.031$ & $+0.032$ \\
\midrule
\multicolumn{9}{l}{\textbf{POS class unknown, vectors averaged}} \\
\multirow{2}{*}{MEN, all pairs} & \multirow{2}{*}{3,000} & 1 & \textbf{0.727} & 0.709 & $-0.020$ & $-0.018$ & $-0.017$ & $-0.015$ \\
 &  & 5 & \textbf{0.740} & 0.726 & $-0.014$ & $-0.014$ & $-0.014$ & $-0.015$ \\
\multirow{2}{*}{WS-353, all pairs} & \multirow{2}{*}{353} & 1 & \textbf{0.652} & 0.644 & $-0.014$ & $-0.008$ & $-0.007$ & $+0.001$ \\
 &  & 5 & \textbf{0.680} & 0.675 & $-0.011$ & $-0.005$ & $+0.002$ & $-0.002$ \\
\bottomrule
\end{tabular*}
\end{center}
\end{table}

POS tags help verbs. SimVerb-3500 rises by 0.016 to 0.033 at every dimension from 50 to 300 in both windows (Table~\ref{tab:tags}), YP-130 rises by 0.057 to 0.106 on its 130 pairs, and SimLex-999 verbs rises in seven of eight settings. The least ambiguous third of the SimVerb-3500 pairs, whose two verbs occur as verbs in at least 86 percent of their occurrences, gains too: 0.027 and 0.015 at 100 dimensions in the two windows. Nouns gain less. The main noun benchmark, SimLex-999 nouns, stays within 0.007 of the original at every dimension. MEN nouns, RG-65 and RW nouns are higher with tags at every dimension, MEN nouns by at most 0.011, RG-65 by 0.008 to 0.057 on its 65 pairs and RW nouns by 0.002 to 0.030, while WS-353 nouns and MTurk-771 nouns are lower in most settings, MTurk-771 nouns by 0.010 to 0.023 at 200 and 300 dimensions. MC-30, with 30 pairs, gains 0.083 to 0.122 at 50 and 100 dimensions but little at 200 and 300 ($-0.005$ to $+0.039$). Adjectives behave like verbs: SimLex-999 adjectives gains in every setting, by 0.059 and 0.052 at 100 dimensions and by 0.006 to 0.040 at the others. Window size has its usual effect \cite{levy2015improving} in Table~\ref{tab:tags}: SimVerb-3500, SimLex-999 and YP-130 score higher in window 1, and MEN, WS-353 and MC-30 in window 5, for verbs and nouns alike.

\textbf{The vectors must stay separate.} When the POS class is unknown and the tagged model represents a word by the count-weighted average of its rows, it falls 0.014 to 0.020 below the original on MEN with all pairs at every dimension and window, and 0.008 and 0.005 below on WS-353 at 100 dimensions. On SimVerb-3500 at 100 dimensions, the tagged table gains 0.033 and 0.021 over the original in the two windows and the averaged vectors keep 0.008 of it in both; at 300 dimensions they keep at most 0.001 of the 0.029 and 0.024 gain, scoring 0.337 and 0.282 against 0.341 and 0.281 for the original table. Averaging them back thus loses the verb gain and, on MEN, does not even match a table with one vector per word, so the vectors of a word must stay separate and be chosen by part of speech, as they are in the rest of this paper.

\section{Which Context a Part of Speech Needs}\label{sec:context}

Section~\ref{sec:tags} changed how tokens are written; this section changes which tokens a POS class sees as context. In the full corpus a verb's context mixes every part of speech. To see which of them a verb needs, we start from the verb-only corpus, where verbs see only other verbs, and add one POS class back: each line's adverbs, the class that modifies verbs. For nouns we start from the noun-only corpus and add each line's adjectives. Each addition has a baseline that adds as many random tokens of the same line (Table~\ref{tab:context}), so a gain over the baseline comes from the added class and not from the extra text.

\begin{table}[!t]
\begin{center}
\caption{Adding a line's adverbs to the verb-only corpus and its adjectives to the noun-only corpus, against adding as many random words from the same lines. Spearman correlation, means over 3 seeds. (a) Scores at 100 dimensions; the highest of the three in each row is in bold, and $\Delta$ is the gain over the random baseline, verb+adverb minus verb+random or noun+adjective minus noun+random. (b) Differences between corpora on SimVerb-3500 and SimLex-999 nouns at 50 to 300 dimensions. $\Delta$ is computed before rounding.\label{tab:context}}
\small
\setlength{\tabcolsep}{3pt}
\renewcommand{\arraystretch}{1.0}
\begin{tabular*}{\linewidth}{@{\extracolsep{\fill}}lccccc@{}}
\toprule
\multicolumn{6}{l}{\textbf{(a) Scores at 100 dimensions}} \\
\midrule
Benchmark & $w$ & verb-only & verb+random & verb+adverb & $\Delta$ \\
\midrule
\multirow{2}{*}{SimVerb-3500} & 1 & 0.289 & 0.288 & \textbf{0.327} & $+0.039$ \\
 & 5 & 0.269 & 0.259 & \textbf{0.292} & $+0.032$ \\
\multirow{2}{*}{SimLex-999 verbs} & 1 & 0.182 & 0.172 & \textbf{0.282} & $+0.111$ \\
 & 5 & 0.154 & 0.134 & \textbf{0.206} & $+0.072$ \\
\multirow{2}{*}{YP-130} & 1 & 0.640 & 0.635 & \textbf{0.647} & $+0.012$ \\
 & 5 & 0.622 & 0.627 & \textbf{0.646} & $+0.019$ \\
\midrule
Benchmark & $w$ & noun-only & noun+random & noun+adjective & $\Delta$ \\
\midrule
\multirow{2}{*}{SimLex-999 nouns} & 1 & 0.400 & 0.404 & \textbf{0.409} & $+0.005$ \\
 & 5 & 0.375 & 0.382 & \textbf{0.384} & $+0.002$ \\
\multirow{2}{*}{RW nouns} & 1 & 0.327 & 0.331 & \textbf{0.343} & $+0.012$ \\
 & 5 & 0.345 & 0.344 & \textbf{0.361} & $+0.017$ \\
\multirow{2}{*}{MC-30} & 1 & \textbf{0.803} & 0.794 & 0.783 & $-0.011$ \\
 & 5 & \textbf{0.859} & 0.843 & 0.855 & $+0.013$ \\
\multirow{2}{*}{RG-65} & 1 & 0.818 & \textbf{0.827} & 0.826 & $-0.001$ \\
 & 5 & 0.823 & 0.827 & \textbf{0.835} & $+0.008$ \\
\multirow{2}{*}{MEN nouns} & 1 & \textbf{0.757} & 0.756 & 0.754 & $-0.002$ \\
 & 5 & \textbf{0.762} & 0.760 & 0.757 & $-0.002$ \\
\multirow{2}{*}{WS-353 nouns} & 1 & 0.656 & 0.667 & \textbf{0.668} & $+0.001$ \\
 & 5 & 0.698 & \textbf{0.706} & \textbf{0.706} & $0.000$ \\
\multirow{2}{*}{MTurk-771 nouns} & 1 & 0.696 & 0.701 & \textbf{0.710} & $+0.008$ \\
 & 5 & 0.697 & 0.696 & \textbf{0.705} & $+0.009$ \\
\bottomrule
\end{tabular*}

\vspace{6pt}
\begin{tabular*}{\linewidth}{@{\extracolsep{\fill}}lccccc@{}}
\toprule
\multicolumn{6}{l}{\textbf{(b) Differences at 50 to 300 dimensions}} \\
\midrule
Difference & $w$ & 50 & 100 & 200 & 300 \\
\midrule
\multicolumn{6}{l}{\textit{SimVerb-3500}} \\
\multirow{2}{*}{(verb+adverb) $-$ (verb+random)} & 1 & $+0.034$ & $+0.039$ & $+0.043$ & $+0.046$ \\
 & 5 & $+0.027$ & $+0.032$ & $+0.037$ & $+0.036$ \\
\multirow{2}{*}{(verb+adverb) $-$ (verb-only)} & 1 & $+0.025$ & $+0.038$ & $+0.044$ & $+0.048$ \\
 & 5 & $+0.018$ & $+0.023$ & $+0.034$ & $+0.036$ \\
\multirow{2}{*}{(verb+random) $-$ (verb-only)} & 1 & $-0.009$ & $-0.001$ & $+0.001$ & $+0.002$ \\
 & 5 & $-0.009$ & $-0.009$ & $-0.002$ & $-0.001$ \\
\midrule
\multicolumn{6}{l}{\textit{SimLex-999 nouns}} \\
\multirow{2}{*}{(noun+adjective) $-$ (noun+random)} & 1 & $+0.012$ & $+0.005$ & $+0.008$ & $+0.007$ \\
 & 5 & $+0.003$ & $+0.002$ & $+0.009$ & $+0.010$ \\
\bottomrule
\end{tabular*}
\end{center}
\end{table}

\textbf{Adverbs for verbs.} At every dimension from 50 to 300 and in both windows, verb+adverb beats both verb-only and verb+random on SimVerb-3500 and SimLex-999 verbs. SimVerb-3500 labels each pair with its lexical relation \cite{gerz2016simverb}: synonyms, antonyms, hypernym and hyponym, cohyponyms, or none. From 100 to 300 dimensions, verb+adverb leads both baselines on all five in both windows; at 50 dimensions it falls 0.004 and 0.007 below verb-only on cohyponyms.

\textbf{Adjectives for nouns.} Adjectives help nouns much less than adverbs help verbs. Noun+adjective is above noun+random on SimLex-999 nouns at every dimension, by 0.002 to 0.012 (Table~\ref{tab:context}(b)), and at 100 dimensions in nine of the fourteen settings of the seven noun benchmarks, by at most 0.017 (Table~\ref{tab:context}(a)). Over the four dimensions the differences stay small; RW nouns in window 5 gains 0.014 to 0.021 at every dimension. So adjectives give nouns a small gain, well below what adverbs give verbs. The two comparisons are not on equal footing, though: a verb-only line has 4.93 tokens and a noun-only line 13.14, so the same number of added words changes a verb's contexts more than a noun's, and each addition is judged only against its own random-word baseline; the difference between the two gains cannot be attributed to part of speech alone.

\textbf{Verb tables reach nearly their full score by 200 dimensions.} From 100 to 200 dimensions, SimVerb-3500 rises by 0.014 to 0.026 for verb-only and verb+adverb, and from 200 to 300 by only 0.004 to 0.008, against 0.018 to 0.026 for the original and tagged tables. Their lines are short, 4.93 tokens for verb-only against 13.14 for noun-only and 30.01 for the full corpus (Table~\ref{tab:corpora}), so a window of 5 usually spans a whole verb line and the two windows differ less there.

\textbf{Why we train one table.} A POS class trained alone loses much of the corpus and the context words other POS classes supply. Adverbs can make a separate verb table competitive with the tagged table on SimVerb-3500: at 100 dimensions verb+adverb scores 0.327 and 0.292 (Table~\ref{tab:context}) against 0.303 and 0.250 for the tagged table (Table~\ref{tab:tags}), and at 300 dimensions the two are within 0.02 of each other. But a set of per-POS tables, verb+adverb at 300 dimensions, noun-only at 300 and adjective+adverb-only at 200 with each word routed by its part of speech (2,372 MB), trails the tagged table by 0.026 on SimLex-999 nouns and 0.035 on RW nouns at $w{=}1$, and puts verbs and nouns in separate spaces, where they cannot be compared. Training with fewer dimensions does not help either: every single-POS table scores highest at 300 dimensions on SimVerb-3500, SimLex-999 nouns or SimLex-999 adjectives, except adjective+adverb-only at $w{=}1$, where 300 is 0.001 below 200. We therefore train one tagged table and reduce its dimensions only after training (Sections~\ref{sec:geometry} and \ref{sec:method}). The context effects of this section are then inside that table: Section~\ref{sec:geometry} asks whether the parts of speech also occupy directions of their own and need different numbers of them, and Section~\ref{sec:method} what each keeps when the table must lose dimensions.
\section{Subspaces and Dimensions by Part of Speech}\label{sec:geometry}

We now ask whether the parts of speech occupy their own directions in the trained space, how many dimensions each needs, and what separating them costs. We split the rows of the 300-dimensional tagged table into six POS classes: verbs (VERB and AUX rows), nouns, proper nouns, adjectives, adverbs and other words. Proper nouns are a class of their own here: they have their own POS tag and fill 72.7 percent of the rows, whereas the noun corpora and benchmarks of Sections~\ref{sec:tags} and \ref{sec:context} count them as nouns. Other words take every remaining POS tag, from determiners and pronouns to numerals and symbols. They are not just function words, which are frequent: 89.7 percent of their 52,544 rows occur fewer than 100 times.

\subsection{Per-POS PCA and count weighting}\label{sec:weighting}

Per-POS PCA fits its own mean and basis to each part of speech, as local PCA does \cite{kambhatla1997dimension} and as PCA-based feedforward designs do \cite{kuo2019interpretable}. We scale every row $x_i$ to unit length, $u_i = x_i / \|x_i\|$, and weight it by its training count $n_i$ in the mean and covariance of its POS class $c$,
\begin{equation}
\mu_c = \frac{\sum_{i \in c} n_i u_i}{\sum_{i \in c} n_i}, \qquad
C_c = \frac{\sum_{i \in c} n_i (u_i - \mu_c)(u_i - \mu_c)^{\top}}{\sum_{i \in c} n_i}.
\end{equation}
The eigenvectors of $C_c$, ordered by decreasing eigenvalue, form the columns of $U_c$, the PCA basis of the POS class. A row keeps its first $k$ coefficients $z$ and is decoded into the shared 300-dimensional space as $\hat{u} = \mu_c + U_{c,k} z$. At $k = 300$ the transform is a centered rotation, so decoding recovers every unit-length row up to floating-point error.

Count weighting lets each word count as often as it occurs, instead of once as in an unweighted PCA. We use it throughout because the paper is about the similarity of common words; its cost to verb-to-noun retrieval is measured in Section~\ref{sec:crosspos}.

\subsection{POS bases against global and band bases}\label{sec:subspaces}

\begin{figure}[!t]
\centering
\includegraphics[width=0.8\linewidth]{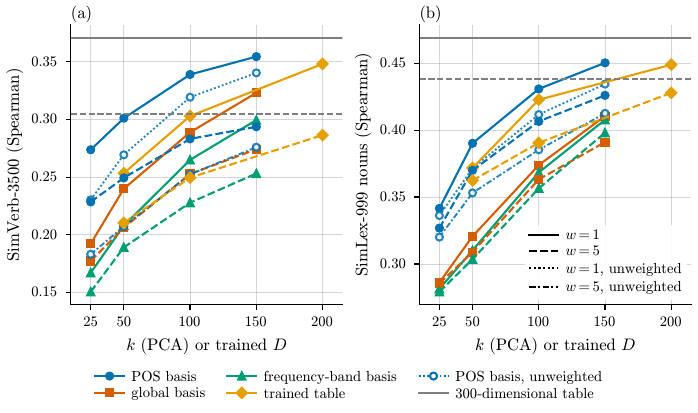}
\caption{Spearman correlation (3-seed means) against the number of dimensions $k$ for the POS bases (per-POS PCA), the global basis, the band bases and a table trained at $D = k$ dimensions, on (a) SimVerb-3500 and (b) SimLex-999 nouns. Bases are count-weighted unless marked unweighted. Solid lines are $w{=}1$ and dashed lines $w{=}5$; gray horizontal lines mark the 300-dimensional table in each window. Open circles (dotted for $w{=}1$, dash-dotted for $w{=}5$) show the POS bases without count weighting (Section~\ref{sec:weighting}). Table~\ref{tab:bases} lists the scores of all three.}\label{fig:subspaces}
\end{figure}

\begin{table}[!t]
\begin{center}
\caption{The POS bases against the global basis and the band bases at $k = 25$ to $150$ dimensions, all count-weighted (Spearman correlation, means over 3 seeds). In each column the highest of the three rows is in bold; the scores of the full 300-dimensional table ($w{=}1$\,/\,$w{=}5$) follow each benchmark name.\label{tab:bases}}
\small
\setlength{\tabcolsep}{3pt}
\renewcommand{\arraystretch}{1.0}
\begin{tabular*}{\linewidth}{@{\extracolsep{\fill}}lcccccccc@{}}
\toprule
 & \multicolumn{2}{c}{$k = 25$} & \multicolumn{2}{c}{$k = 50$} & \multicolumn{2}{c}{$k = 100$} & \multicolumn{2}{c}{$k = 150$} \\
\cmidrule(lr){2-3}\cmidrule(lr){4-5}\cmidrule(lr){6-7}\cmidrule(lr){8-9}
Basis & $w{=}1$ & $w{=}5$ & $w{=}1$ & $w{=}5$ & $w{=}1$ & $w{=}5$ & $w{=}1$ & $w{=}5$ \\
\midrule
\multicolumn{9}{l}{\textbf{SimVerb-3500} (full 300-dimensional table 0.371\,/\,0.305)} \\
global & 0.192 & 0.177 & 0.240 & 0.206 & 0.289 & 0.253 & 0.323 & 0.274 \\
frequency band & 0.167 & 0.151 & 0.208 & 0.189 & 0.265 & 0.228 & 0.300 & 0.253 \\
POS & \textbf{0.274} & \textbf{0.229} & \textbf{0.301} & \textbf{0.249} & \textbf{0.339} & \textbf{0.283} & \textbf{0.355} & \textbf{0.294} \\
\midrule
\multicolumn{9}{l}{\textbf{SimLex-999 verbs} (full 300-dimensional table 0.324\,/\,0.246)} \\
global & 0.057 & 0.041 & 0.149 & 0.113 & 0.242 & 0.192 & 0.279 & 0.222 \\
frequency band & 0.043 & 0.015 & 0.117 & 0.105 & 0.217 & 0.171 & 0.252 & 0.202 \\
POS & \textbf{0.250} & \textbf{0.199} & \textbf{0.272} & \textbf{0.211} & \textbf{0.290} & \textbf{0.232} & \textbf{0.309} & \textbf{0.241} \\
\midrule
\multicolumn{9}{l}{\textbf{SimLex-999 adjectives} (full 300-dimensional table 0.555\,/\,0.502)} \\
global & 0.269 & 0.259 & 0.346 & 0.319 & 0.395 & 0.377 & 0.457 & 0.450 \\
frequency band & 0.180 & 0.215 & 0.240 & 0.271 & 0.351 & 0.360 & 0.432 & 0.412 \\
POS & \textbf{0.295} & \textbf{0.344} & \textbf{0.465} & \textbf{0.422} & \textbf{0.517} & \textbf{0.482} & \textbf{0.544} & \textbf{0.492} \\
\midrule
\multicolumn{9}{l}{\textbf{SimLex-999 nouns} (full 300-dimensional table 0.469\,/\,0.439)} \\
global & 0.286 & 0.285 & 0.321 & 0.309 & 0.374 & 0.363 & 0.411 & 0.391 \\
frequency band & 0.281 & 0.279 & 0.310 & 0.304 & 0.368 & 0.357 & 0.408 & 0.398 \\
POS & \textbf{0.342} & \textbf{0.327} & \textbf{0.390} & \textbf{0.370} & \textbf{0.431} & \textbf{0.407} & \textbf{0.450} & \textbf{0.426} \\
\midrule
\multicolumn{9}{l}{\textbf{YP-130} (full 300-dimensional table 0.645\,/\,0.637)} \\
global & 0.428 & 0.451 & 0.499 & 0.481 & 0.566 & 0.548 & \textbf{0.605} & 0.585 \\
frequency band & 0.384 & 0.356 & 0.439 & 0.429 & 0.518 & 0.530 & 0.563 & 0.566 \\
POS & \textbf{0.483} & \textbf{0.461} & \textbf{0.518} & \textbf{0.511} & \textbf{0.577} & \textbf{0.584} & 0.603 & \textbf{0.622} \\
\bottomrule
\end{tabular*}
\end{center}
\end{table}

Section~\ref{sec:weighting} gave each part of speech its own PCA basis. This section tests whether that basis describes a POS class's rows better than one basis for the whole table. We keep $k$ directions of each row in three ways. The POS bases are the per-POS PCA of Section~\ref{sec:weighting}, one basis per class. The global basis is one PCA fitted to all rows. The band bases come from one PCA per band of training count (split at $10^2$, $10^3$, $10^4$ and $10^5$). They are also several bases instead of one, but group the rows by frequency, so the two words of a benchmark pair, which share a part of speech, are truncated in one basis under the POS bases and often in two under the band bases; the cleaner test of whether a POS class is better described by its own basis is the comparison with the global basis. The decoded tables are scored on the benchmarks at the same $k$, and all bases are count-weighted unless we say otherwise. If a POS class is better described by its own basis, the POS bases should score higher.

It does, for verbs, adjectives and nouns alike. From 25 to 150 dimensions, the POS bases score higher than both alternatives on all five benchmarks in both windows (Figure~\ref{fig:subspaces}, Table~\ref{tab:bases}). The one exception is YP-130 at $k = 150$ in window 1, where the POS and global bases are within 0.002 of each other. Over the band bases the lead is 0.041 to 0.107 on SimVerb-3500 and 0.028 to 0.080 on SimLex-999 nouns (Table~\ref{tab:bases}), but the band bases also split the two words of a pair across bases, so this lead is not a clean measure of what the part of speech adds. At $k = 100$ the POS bases also score above a table trained at 100 dimensions on both SimVerb-3500 and SimLex-999 nouns (Figure~\ref{fig:subspaces}).

Count weighting might be doing part of this work, so we refitted all three without it, each row counting once. For verbs and adjectives the answer does not change: the POS bases still score higher than the global basis and the band bases on SimVerb-3500 and SimLex-999 adjectives at every $k$ in both windows. For nouns it does: with unweighted bases, the POS bases score 0.412 and 0.385 on SimLex-999 nouns at 100 dimensions against 0.406 and 0.378 for the global basis, within 0.01 in both windows. So most of the noun gain in Table~\ref{tab:bases} at that size comes from count weighting, not from the basis itself.

Beyond the benchmark scores, we also look at a word's nearest neighbors after reduction and ask whether they still belong to its semantic category. The queries are the 1,669 verbs and 4,927 nouns with training count at least $10^4$, and the candidates are rows of the same part of speech with count at least $10^3$. Both must have a WordNet sense of that part of speech. A word's semantic category is the lexicographer file of its first WordNet synset for that part of speech \cite{fellbaum1998wordnet,tsvetkov2015evaluation}, which gives 15 verb and 26 noun categories. Purity@10 is the share of a query's 10 nearest candidates that fall in its semantic category.

At $k = 100$ the POS bases keep more neighbors in the query's category than the global basis, for verbs and nouns in both windows: purity@10 is 0.367 and 0.340 for verbs and 0.484 and 0.441 for nouns with the POS bases, against 0.344 and 0.326 and 0.461 and 0.424 with the global basis, higher by 0.014 to 0.023; the full table scores 0.372 and 0.343 and 0.477 and 0.426. Compared with the full 300-dimensional table, the POS bases are slightly lower for verbs and higher for nouns, while the global basis is lower for both. Table~\ref{tab:nbexamples} gives examples. Two neighbors of \textit{know}, \textit{nickname} and \textit{dub}, come from lemmatization, which sends \textit{known as} to the row of \textit{know}.

\begin{table}[!t]
\begin{center}
\caption{Five nearest neighbors of one query in each of four verb and four noun WordNet lexicographer files, the semantic categories used for purity@10 (verbs above the line, nouns below; seed 1, $w{=}1$). Each query is the most frequent query of its file; for possession this is the auxiliary \textit{have}. $\checkmark$: the neighbor is in the query's file; the counts in the last row cover these eight queries only, whereas purity@10 in the text is over all queries. The file comes from the first WordNet synset, which misleads for polysemous words: \textit{follow} is filed under motion, but its neighbors in every system are stative, change or creation verbs.}\label{tab:nbexamples}
\small
\setlength{\tabcolsep}{3pt}
\renewcommand{\arraystretch}{1.1}
\setlength{\extrarowheight}{2pt}
\begin{tabular}{@{}>{\raggedright\arraybackslash}p{3.0cm}>{\raggedright\arraybackslash}p{4.25cm}>{\raggedright\arraybackslash}p{4.25cm}>{\raggedright\arraybackslash}p{4.25cm}@{}}
\toprule
WordNet file, query & full table, 300 & global basis, $k{=}100$ & POS bases, $k{=}100$ \\
\midrule
motion:\newline \textit{follow} & precede, coincide, culminate, result, prompt & coincide, precede, culminate, result, end & precede, coincide, culminate, result, ensue \\
communication:\newline \textit{call} & term\,$\checkmark$, dub\,$\checkmark$, label\,$\checkmark$, refer\,$\checkmark$, title\,$\checkmark$ & term\,$\checkmark$, dub\,$\checkmark$, label\,$\checkmark$, refer\,$\checkmark$, title\,$\checkmark$ & term\,$\checkmark$, dub\,$\checkmark$, label\,$\checkmark$, title\,$\checkmark$, refer\,$\checkmark$ \\
cognition:\newline \textit{know} & remember\,$\checkmark$, nickname, dub, call, refer & nickname, dub, call, refer, remember\,$\checkmark$ & nickname, remember\,$\checkmark$, dub, repute\,$\checkmark$, acquaint \\
possession:\newline \textit{have} & be, will, need, want, possess & possess, need, expect, will, want & possess, be, need, will, poise \\
\midrule
animal:\newline \textit{male} & female\,$\checkmark$, mating, copulation, juvenile, adult & female\,$\checkmark$, juvenile, adult, individual, chick\,$\checkmark$ & female\,$\checkmark$, adult, nestling\,$\checkmark$, juvenile, chick\,$\checkmark$ \\
artifact:\newline \textit{system} & subsystem, network, scheme, mechanism, technology & subsystem, device\,$\checkmark$, technology, mechanism, automation & subsystem, circuitry\,$\checkmark$, scheme, setup\,$\checkmark$, network \\
food:\newline \textit{wine} & brandy\,$\checkmark$, grape\,$\checkmark$, winery, cider\,$\checkmark$, beer\,$\checkmark$ & cheese\,$\checkmark$, grape\,$\checkmark$, beer\,$\checkmark$, brandy\,$\checkmark$, strawberry\,$\checkmark$ & beer\,$\checkmark$, cider\,$\checkmark$, whisky\,$\checkmark$, coffee\,$\checkmark$, brandy\,$\checkmark$ \\
person:\newline \textit{member} & representative\,$\checkmark$, membership, chairperson\,$\checkmark$, chairman\,$\checkmark$, leader\,$\checkmark$ & representative\,$\checkmark$, membership, delegate\,$\checkmark$, leader\,$\checkmark$, alumnus\,$\checkmark$ & representative\,$\checkmark$, membership, chairperson\,$\checkmark$, fellow\,$\checkmark$, alumnus\,$\checkmark$ \\
\midrule
$\checkmark$ out of 40 & 15 & 18 & 21 \\
\bottomrule
\end{tabular}
\end{center}
\end{table}

\begin{figure}[!t]
\centering
\includegraphics[width=0.8\linewidth]{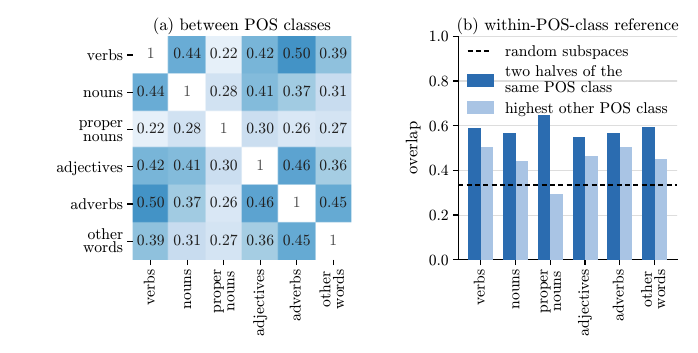}
\caption{How much the subspaces of the parts of speech overlap, measured as the mean squared cosine of the principal angles between the top 100 count-weighted PCA directions of two POS classes, averaged over 3 seeds and 2 windows. Two random 100-dimensional subspaces of the 300-dimensional space overlap 1/3 on average (dashed line in (b)). (a) Overlap between POS classes; darker cells overlap more, and the unshaded diagonal is 1 by definition. The split-half overlap in (b) is the reference within a POS class. (b) For each POS class, the overlap of two random halves of the POS class, each fitted separately, next to its overlap with the other POS class it overlaps most. Split-half overlap is 0.543 to 0.619 in window 1 and 0.553 to 0.679 in window 5, and overlap between POS classes 0.216 to 0.471 and 0.233 to 0.534, respectively.}\label{fig:overlap}
\end{figure}

The scores show that a basis fitted to a POS class describes it better. Whether the class also has directions of its own is a different question, which Figure~\ref{fig:overlap} answers by measuring how much the top 100 count-weighted directions of two POS classes overlap. As a baseline, two random 100-dimensional subspaces of the 300-dimensional space overlap 1/3 on average. Averaged over the two windows, two random halves of the same POS class, each fitted separately, overlap 0.548 to 0.649, and two different POS classes 0.224 to 0.503 (the caption of Figure~\ref{fig:overlap} gives each window). In every POS class, the two halves overlap more than the POS class overlaps with any other. The margin is largest for proper nouns, about 0.35; it is 0.12 for nouns and 0.14 for other words, and 0.065 to 0.088 for verbs, adjectives and adverbs, which are closest to one another. So every POS class has directions of its own: proper nouns by far the most, verbs, adjectives and adverbs the fewest.

\subsection{How many dimensions each POS class needs}\label{sec:intrinsic}

\begin{figure}[!t]
\centering
\includegraphics[width=0.8\linewidth]{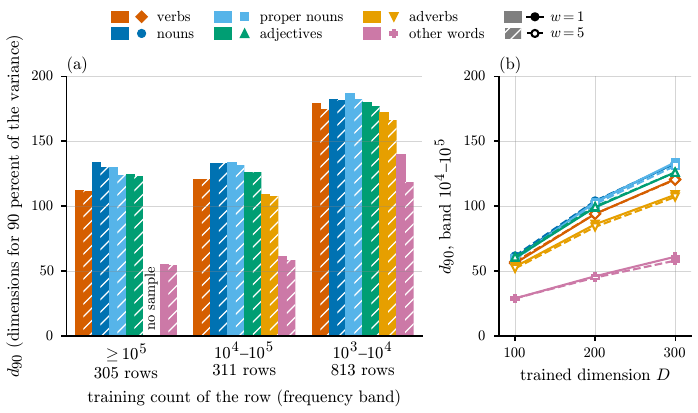}
\caption{Dimensions needed by part of speech. (a) $d_{90}$ (Table~\ref{tab:dims}; 3-seed means) on equal-size samples from one frequency band per POS class; the number of rows per POS class is given under each band, and the top band has no adverb sample. Plain bars are $w{=}1$ and hatched bars $w{=}5$. Compare POS classes only within a band, because the sample size differs between bands. (b) $d_{90}$ of the $10^4$ to $10^5$ samples in tables trained at 100, 200 and 300 dimensions (3-seed means); solid lines are $w{=}1$ and dashed lines $w{=}5$.}\label{fig:dims}
\end{figure}

We measure how many dimensions a POS class needs as $d_{90}$, the number of principal components holding 90 percent of a sample's variance in an unweighted PCA. Among words seen 100 to 1,000 times, every POS class needs about the same number of dimensions: their $d_{90}$ values all fall between 192 and 204. So we measure on words seen at least 1,000 times, one frequency band at a time. Within a band we drop POS classes with fewer than 300 rows and draw from each of the others as many rows as the smallest one has, use the same rows for every seed and window, and run the PCA on unit-length rows centered on the sample mean (Table~\ref{tab:dims}, Figure~\ref{fig:dims}). The PCA is unweighted and the samples have equal size so that no class is favored by its row count or by the counts of its words; the count-weighted bases of Section~\ref{sec:weighting} serve a different purpose, a small error on the words that occur most. The samples are used only here and for the per-POS lengths of Section~\ref{sec:method-def}; the POS bases are fitted to all rows of each class, and the benchmarks are scored on the full decoded table.

Nouns need more dimensions than verbs, by 12 to 22 directions at $d_{90}$ in the two top bands and by 3 to 7 in the $10^3$ to $10^4$ band, and verbs more than other words, by 39 to 62, in every band and window (Table~\ref{tab:dims}). The same order holds at 95 and 99 percent of the variance ($d_{95}$ and $d_{99}$ in Table~\ref{tab:dims}) and, in the $10^4$ to $10^5$ band, on tables trained at 100 and 200 dimensions (Figure~\ref{fig:dims}(b)). Proper nouns are close to nouns. Neighbor retention gives the same order in the top band, where $k_{80}$ is 125 for verbs, 175 for nouns and 50 for other words, but not between $10^3$ and $10^4$, where other words need more than verbs, 250 and 200 against 175 and 175 (Table~\ref{tab:dims}).

\begin{table}[!t]
\begin{center}
\caption{Dimensions needed by each part of speech, on equal-size samples of $n$ rows per POS class from one frequency band (the top band has no adverb sample). $d_{90}$, $d_{95}$ and $d_{99}$: principal directions that keep 90, 95 and 99 percent of the sample's variance in an unweighted PCA of the 300-dimensional table (3-seed means). $k_{80}$: the smallest $k$ in 10, 25, 50, 75, \ldots, 300 at which a sample word keeps 80 percent of its 10 nearest neighbors, among rows of its POS class with count $\ge 10^{3}$, after reduction to $k$ count-weighted directions of its class (median over 3 seeds). Cells give $w{=}1$ / $w{=}5$. Within a band every POS class has the same $n$; $n$ differs between bands, so compare values only within a band.}\label{tab:dims}
\small
\setlength{\tabcolsep}{3pt}
\renewcommand{\arraystretch}{1.15}
\begin{tabular*}{\linewidth}{@{\extracolsep{\fill}}lcccc@{}}
\toprule
 & \multicolumn{4}{c}{$w{=}1$\,/\,$w{=}5$} \\
POS class & $d_{90}$ & $d_{95}$ & $d_{99}$ & $k_{80}$ \\
\midrule
\multicolumn{5}{l}{\textbf{count $\ge 10^{5}$, $n = 305$}} \\
verbs & 112.0\,/\,111.0 & 145.3\,/\,144.7 & 205.0\,/\,204.0 & 125\,/\,125 \\
nouns & 134.0\,/\,130.0 & 165.0\,/\,162.0 & 219.3\,/\,217.0 & 175\,/\,175 \\
proper nouns & 129.7\,/\,123.3 & 160.7\,/\,154.7 & 213.3\,/\,209.7 & 175\,/\,175 \\
adjectives & 124.3\,/\,123.0 & 156.0\,/\,155.7 & 212.0\,/\,212.0 & 125\,/\,150 \\
other words & 55.0\,/\,54.0 & 82.0\,/\,81.0 & 147.0\,/\,147.0 & 50\,/\,50 \\
\midrule
\multicolumn{5}{l}{\textbf{count $10^{4}$ to $10^{5}$, $n = 311$}} \\
verbs & 120.7\,/\,120.3 & 154.7\,/\,154.0 & 213.7\,/\,213.3 & 175\,/\,150 \\
nouns & 133.0\,/\,132.7 & 166.0\,/\,165.3 & 220.7\,/\,220.7 & 175\,/\,175 \\
proper nouns & 133.7\,/\,131.0 & 165.7\,/\,163.0 & 220.3\,/\,218.0 & 225\,/\,200 \\
adjectives & 126.0\,/\,126.0 & 159.0\,/\,158.0 & 215.7\,/\,214.3 & 175\,/\,150 \\
adverbs & 108.7\,/\,107.0 & 143.0\,/\,140.7 & 204.7\,/\,203.7 & 125\,/\,125 \\
other words & 61.0\,/\,58.0 & 94.0\,/\,90.0 & 169.7\,/\,163.7 & 100\,/\,100 \\
\midrule
\multicolumn{5}{l}{\textbf{count $10^{3}$ to $10^{4}$, $n = 813$}} \\
verbs & 179.0\,/\,174.3 & 220.0\,/\,216.0 & 274.7\,/\,272.3 & 175\,/\,175 \\
nouns & 182.3\,/\,181.3 & 222.3\,/\,221.0 & 275.0\,/\,274.0 & 200\,/\,200 \\
proper nouns & 187.0\,/\,182.3 & 225.0\,/\,221.0 & 276.0\,/\,273.0 & 225\,/\,200 \\
adjectives & 180.0\,/\,176.3 & 220.7\,/\,217.0 & 275.0\,/\,273.0 & 200\,/\,175 \\
adverbs & 172.0\,/\,166.0 & 215.0\,/\,209.3 & 273.0\,/\,270.0 & 150\,/\,150 \\
other words & 140.0\,/\,118.0 & 190.0\,/\,169.7 & 261.0\,/\,250.0 & 250\,/\,200 \\
\bottomrule
\end{tabular*}
\end{center}
\end{table}

\subsection{The cost to cross-POS relations}\label{sec:crosspos}

Separate bases may pull the POS classes apart. We test this on verb-to-noun retrieval, the one cross-POS relation we measured (Section~\ref{sec:evaluation}): 5,646 verb rows query 129,677 noun rows. At $k = 100$ in window 1, MRR falls from 0.479 for the full table to 0.245 for the POS bases, while the count-weighted global basis keeps 0.401. Both fall further at $k = 25$, and window 5 shows the same pattern (Table~\ref{tab:retrieval}).

\begin{table}[!t]
\begin{center}
\caption{Verb-to-derived-noun retrieval after truncation to $k$ directions and decoding (MRR, 3-seed means, $w{=}1$\,/\,$w{=}5$). The full 300-dimensional table scores 0.479\,/\,0.569. In the unweighted rows, both bases are fitted without count weighting.}\label{tab:retrieval}
\small
\setlength{\tabcolsep}{3pt}
\renewcommand{\arraystretch}{1.15}
\begin{tabular*}{\linewidth}{@{\extracolsep{\fill}}lcc@{}}
\toprule
Basis & $k = 25$ & $k = 100$ \\
\midrule
POS, count-weighted & 0.025\,/\,0.067 & 0.245\,/\,0.359 \\
global, count-weighted & 0.120\,/\,0.241 & 0.401\,/\,0.495 \\
POS, unweighted & 0.207\,/\,0.286 & 0.395\,/\,0.508 \\
global, unweighted & 0.253\,/\,0.315 & 0.428\,/\,0.521 \\
\bottomrule
\end{tabular*}
\end{center}
\end{table}

We also separated count weighting from the split into POS subspaces to see which of the two causes the drop. On this task the unweighted global basis is the best of the four reduced tables, 0.428 and 0.521 at $k = 100$. Count weighting alone lowers it by 0.027 and 0.026, the split alone, with both bases unweighted, by 0.033 and 0.013, and the two together by 0.183 and 0.162 (Table~\ref{tab:retrieval}), so the cost comes mostly from combining them. These scores are on decoded vectors: decoding puts the POS classes back in one space but does not restore the relations between them that truncation removed. An application that needs verb-to-noun relations can therefore keep the POS bases and drop count weighting, or keep more directions for common words, as the allocation of Section~\ref{sec:method} does, under which, from a third of the table's storage on, the POS bases keep verb-to-noun retrieval within 0.02 of the full table.

In short, the POS bases describe verbs and adjectives better and keep more semantically related neighbors than a global basis, frequent nouns need more dimensions than frequent verbs, and separate count-weighted bases weaken verb-to-noun retrieval.

\section{Spending a Budget of Dimensions}\label{sec:method}

Among frequent words, the POS classes need different numbers of dimensions (Section~\ref{sec:intrinsic}). We now ask what each word keeps when the tagged table must lose dimensions: in which directions it keeps its coefficients, and how many. The budget is a device for seeing this, not a compression method; a table that only needs frequent words can simply drop its rare rows. Part of speech should decide the directions and frequency the number: from a third of the table's storage on, the POS bases with a budget spent by frequency keep the full table's similarity on SimVerb-3500 and SimLex-999 nouns and its verb-to-noun retrieval within 0.02. At those budgets every word seen at least 100 times keeps all 300 dimensions, and the basis then matters only for verb-to-noun retrieval and rare words. Nouns need more dimensions than verbs only among frequent words, so no part of speech needs a length of its own. Against one basis per frequency band, the POS bases keep up to 0.04 more of the verb-to-noun retrieval and lose only on the rare words.

\subsection{The pipeline}\label{sec:method-def}

Figure~\ref{fig:method} shows the pipeline. We train one 300-dimensional table on the tagged corpus (Section~\ref{sec:protocol}) and group its rows into blocks by training count, at the cut points of Section~\ref{sec:subspaces}, so that frequent words can keep more dimensions than rare ones \cite{baevski2018adaptive,chen2018groupreduce,ginart2021mixed}. Every block has its own count-weighted mean and its own number of directions. Given a budget of dimensions for the whole table, a greedy rule hands them out one at a time to the block that gains most from one more direction. The gain is how much closer the decoded vectors of the block come to the original ones, with each word counted as often as it occurs in training, divided by the bytes the extra direction costs. Each row keeps its leading coefficients in float32, as many as its block is allotted, and is decoded with its block mean and the leading columns of its basis into the shared 300-dimensional space. Our storage figures include the bases, the means and a 16-bit norm per row.

The basis follows part of speech. The blocks are formed within each POS class, and all blocks of a class share its count-weighted POS basis $U_c$ (Section~\ref{sec:weighting}), because at one fixed length the POS bases describe verbs, adjectives and nouns better than the global basis or the band bases (Table~\ref{tab:bases}). A block with fewer than 1,000 rows is merged into the next lower band of its class, which leaves 23 blocks. As an ablation we remove the POS split and allocate by frequency only: there is one block per band across all POS classes, five in all, each with its own count-weighted basis (the band bases).

We compare the allocation with the POS bases and with the global basis at uniform lengths of 25 to 150 per row, and with a table trained at 100 dimensions. Two further ablations keep the POS bases and change how the budget is spent. One length per POS class gives every row of a class the same number of directions, set from the dimension needs of Section~\ref{sec:intrinsic} as the number of leading directions that keep 90 or 95 percent of the variance in an unweighted PCA of the samples of the two top bands of Section~\ref{sec:intrinsic} pooled (training count at least $10^4$), so that each class gets one length from all its frequent words, a mean of 163 or 202 directions per row (above the $d_{90}$ values of Table~\ref{tab:dims} for every class with samples in both bands, because the pooled samples have more rows); these lengths make every POS class lose the same share of the sample variance. Allocation by POS class only uses one block per class and the same greedy rule.

\subsection{Results}\label{sec:method-results}

Table~\ref{tab:alloc} scores the allocation with the POS bases at mean budgets of 25 to 150 dimensions per row, the same allocation by frequency only, without the POS split, allocation by POS class only, and the fixed-length tables. Table~\ref{tab:matched} compares the allocation with one length per POS class at the same storage and adds verb-to-noun retrieval and rare-word neighbor retention. The budget is the mean number of dimensions per row over the whole table, and the greedy rule decides where it goes. With the POS bases at a mean of 25, words seen at least 1,000 times keep all 300 directions (299 for proper nouns seen fewer than 10,000 times in window 5), words seen 100 to 999 times keep 78 to 192 on average over seeds and windows, and words seen fewer than 100 times, 86 percent of the rows, keep none: such a row decodes to its block mean, scaled by its norm. At a mean of 100, every word seen at least 100 times keeps all 300 directions, and the lowest band keeps 54 to 88 on average, except other words, which keep 4.

With the POS bases, at a mean of 25 dimensions, 212 MB or 9 percent of the table, SimVerb-3500 is 0.367 and 0.299 and SimLex-999 nouns 0.472 and 0.441, against 0.371 and 0.305 and 0.469 and 0.439 for the full table. From a mean of 100 dimensions on, 829 MB or a third of the table, both benchmarks and MEN with all pairs match the full table, and verb-to-noun MRR is 0.494 and 0.550 against 0.479 and 0.569 (Table~\ref{tab:matched}); in window 1 the MRR is slightly above the full table's, which we do not read as a gain. The same storage spent uniformly does worse: at 100 dimensions per row, the POS bases lose 0.031 and 0.021 on SimVerb-3500 and the global basis 0.081 and 0.051, and a table trained at 100 dimensions is 0.067 and 0.054 below the allocation. Rare words pay: RW nouns falls from 0.373 and 0.375 to 0.289 and 0.260 at a mean of 25 and 0.336 and 0.335 at a mean of 100, where the global basis at a uniform 100 per row, 0.359 and 0.361, and the table trained at 100 dimensions, 0.353 and 0.362, score higher. Removing the POS split changes the three similarity sets by at most 0.002 at every budget and raises RW nouns by up to 0.015 in all but one of the eight cells, where it falls by 0.002 (Table~\ref{tab:alloc}), but lowers verb-to-noun MRR by 0.039 at 829 MB and 0.044 at 1,347 MB in window 1 and by at most 0.006 in window 5 (Table~\ref{tab:matched}).

\begin{table}[!t]
\begin{center}
\caption{Allocation by frequency with the POS bases and, as an ablation, by frequency only, with the band bases (one basis per frequency band, no POS split), at mean budgets of 25 to 150 dimensions per row (212, 418, 829 and 1,240 MB with the POS bases; 212, 417, 826 and 1,235 MB with frequency only), against the full table (2,468 MB), allocation by POS class only at 25 to 150 per row (212, 418, 829 and 1,240 MB), the POS bases or the global basis at a uniform 25 to 150 dimensions per row (212, 418, 829 and 1,240 MB with the POS bases; 210, 416, 827 and 1,239 MB with the global basis) and a table trained at 100 dimensions (823 MB). Spearman correlation, means over 3 seeds, $w{=}1$ / $w{=}5$. MEN is scored on all pairs with the count-weighted average of a word's rows. Scores not more than 0.001 below the full table are in bold.\label{tab:alloc}}
\small
\setlength{\tabcolsep}{2pt}
\renewcommand{\arraystretch}{1.0}
\begin{tabular*}{\linewidth}{@{\extracolsep{\fill}}lcccc@{}}
\toprule
 & SimVerb-3500 & SimLex-999 nouns & MEN, all pairs & RW nouns \\
\midrule
Full table, 300 & 0.371\,/\,0.305 & 0.469\,/\,0.439 & 0.733\,/\,0.747 & 0.373\,/\,0.375 \\
\midrule
\multicolumn{5}{l}{\textit{Allocation by frequency, POS bases, mean dimensions per row}} \\
25 per row & 0.367\,/\,0.299 & \textbf{0.472}\,/\,\textbf{0.441} & 0.731\,/\,0.744 & 0.289\,/\,0.260 \\
50 per row & 0.369\,/\,0.302 & \textbf{0.471}\,/\,\textbf{0.439} & \textbf{0.733}\,/\,\textbf{0.747} & 0.308\,/\,0.301 \\
100 per row & \textbf{0.370}\,/\,\textbf{0.304} & \textbf{0.470}\,/\,\textbf{0.439} & \textbf{0.733}\,/\,\textbf{0.747} & 0.336\,/\,0.335 \\
150 per row & \textbf{0.371}\,/\,\textbf{0.304} & \textbf{0.470}\,/\,\textbf{0.439} & \textbf{0.733}\,/\,\textbf{0.747} & 0.349\,/\,0.349 \\
\midrule
\multicolumn{5}{l}{\textit{Allocation by frequency only (band bases, no POS split), mean dimensions per row}} \\
25 per row & 0.365\,/\,0.298 & \textbf{0.472}\,/\,\textbf{0.439} & \textbf{0.732}\,/\,0.745 & 0.304\,/\,0.270 \\
50 per row & 0.369\,/\,0.302 & \textbf{0.471}\,/\,\textbf{0.439} & \textbf{0.733}\,/\,\textbf{0.747} & 0.323\,/\,0.299 \\
100 per row & \textbf{0.370}\,/\,\textbf{0.304} & \textbf{0.470}\,/\,\textbf{0.439} & \textbf{0.733}\,/\,\textbf{0.747} & 0.349\,/\,0.348 \\
150 per row & \textbf{0.370}\,/\,\textbf{0.304} & \textbf{0.470}\,/\,\textbf{0.439} & \textbf{0.733}\,/\,\textbf{0.747} & 0.359\,/\,0.360 \\
\midrule
\multicolumn{5}{l}{\textit{Allocation by POS class only, mean dimensions per row}} \\
25 per row & 0.353\,/\,0.292 & 0.417\,/\,0.397 & 0.601\,/\,0.634 & 0.302\,/\,0.301 \\
50 per row & 0.363\,/\,0.302 & 0.456\,/\,0.429 & 0.653\,/\,0.681 & 0.337\,/\,0.336 \\
100 per row & 0.369\,/\,\textbf{0.305} & \textbf{0.469}\,/\,\textbf{0.440} & 0.699\,/\,0.718 & 0.349\,/\,0.357 \\
150 per row & \textbf{0.371}\,/\,\textbf{0.305} & \textbf{0.469}\,/\,\textbf{0.441} & 0.720\,/\,0.732 & 0.358\,/\,0.360 \\
\midrule
\multicolumn{5}{l}{\textit{Uniform dimensions per row, POS bases}} \\
25 per row & 0.274\,/\,0.229 & 0.342\,/\,0.327 & 0.538\,/\,0.577 & 0.257\,/\,0.264 \\
50 per row & 0.301\,/\,0.249 & 0.390\,/\,0.370 & 0.602\,/\,0.626 & 0.304\,/\,0.309 \\
100 per row & 0.339\,/\,0.283 & 0.431\,/\,0.407 & 0.659\,/\,0.673 & 0.327\,/\,0.324 \\
150 per row & 0.355\,/\,0.294 & 0.450\,/\,0.426 & 0.699\,/\,0.708 & 0.349\,/\,0.348 \\
\multicolumn{5}{l}{\textit{Uniform dimensions per row, global basis}} \\
25 per row & 0.192\,/\,0.177 & 0.286\,/\,0.285 & 0.510\,/\,0.541 & 0.245\,/\,0.269 \\
50 per row & 0.240\,/\,0.206 & 0.321\,/\,0.309 & 0.596\,/\,0.610 & 0.307\,/\,0.326 \\
100 per row & 0.289\,/\,0.253 & 0.374\,/\,0.363 & 0.651\,/\,0.667 & 0.359\,/\,0.361 \\
150 per row & 0.323\,/\,0.274 & 0.411\,/\,0.391 & 0.685\,/\,0.699 & 0.361\,/\,\textbf{0.377} \\
\midrule
Table trained at 100 & 0.303\,/\,0.250 & 0.423\,/\,0.391 & 0.709\,/\,0.726 & 0.353\,/\,0.362 \\
\bottomrule
\end{tabular*}
\end{center}
\end{table}

\begin{table}[!t]
\begin{center}
\caption{Comparisons within groups of equal storage, with verb-to-noun retrieval (MRR) and rare-word neighbor retention (R1) as defined in Section~\ref{sec:evaluation}: the full table; the allocation by frequency with the POS bases (one basis per POS class) and by frequency only (the band bases, one per frequency band, no POS split) at a mean of 100 dimensions per row, a third of the table, and at a mean of 163, the storage of one length per POS class; one length per POS class at the 90 percent lengths of Section~\ref{sec:method-def}, a mean of 163, against one shared length of 163; and the same pair at the 95 percent lengths, a mean of 202. All tables except frequency only use the POS bases; the fixed lengths give every row of a class, or every row, the same number of dimensions, whereas the allocation splits each class by training count and gives the frequent blocks more dimensions and the rare ones fewer. Spearman correlation, means over 3 seeds, $w{=}1$ / $w{=}5$. Similarity scores not more than 0.001 below the full table are in bold; the allocation rows match the full table on both sets and differ in MRR and R1.\label{tab:matched}}
\small
\setlength{\tabcolsep}{2pt}
\renewcommand{\arraystretch}{1.0}
\begin{tabular*}{\linewidth}{@{\extracolsep{\fill}}lrcccc@{}}
\toprule
 & MB & SimVerb-3500 & SimLex-999 nouns & MRR & R1 \\
\midrule
Full table, 300 & 2,468 & 0.371\,/\,0.305 & 0.469\,/\,0.439 & 0.479\,/\,0.569 & 1.000\,/\,1.000 \\
\midrule
\multicolumn{6}{l}{\textit{About 829 MB, a third of the table, mean 100 per row}} \\
\multicolumn{6}{l}{\textsc{Allocation by frequency}} \\
\quad POS bases & 829 & \textbf{0.370}\,/\,\textbf{0.304} & \textbf{0.470}\,/\,\textbf{0.439} & 0.494\,/\,0.550 & 0.301\,/\,0.324 \\
\quad Frequency only & 826 & \textbf{0.370}\,/\,\textbf{0.304} & \textbf{0.470}\,/\,\textbf{0.439} & 0.455\,/\,0.550 & 0.427\,/\,0.466 \\
\midrule
\multicolumn{6}{l}{\textit{About 1,347 MB, the storage of the per-POS lengths, mean 163 per row}} \\
\multicolumn{6}{l}{\textsc{Allocation by frequency}} \\
\quad POS bases & 1,347 & \textbf{0.371}\,/\,\textbf{0.304} & \textbf{0.470}\,/\,\textbf{0.439} & 0.496\,/\,0.556 & 0.464\,/\,0.488 \\
\quad Frequency only & 1,347 & \textbf{0.370}\,/\,\textbf{0.304} & \textbf{0.470}\,/\,\textbf{0.439} & 0.452\,/\,0.550 & 0.640\,/\,0.690 \\
\multicolumn{6}{l}{\textsc{Fixed lengths}} \\
\quad Per-POS lengths & 1,348 & 0.354\,/\,0.293 & 0.457\,/\,0.430 & 0.365\,/\,0.469 & 0.502\,/\,0.529 \\
\quad Shared length, 163 & 1,346 & 0.357\,/\,0.296 & 0.455\,/\,0.429 & 0.393\,/\,0.492 & 0.501\,/\,0.530 \\
\midrule
\multicolumn{6}{l}{\textit{About 1,666 MB, the per-POS lengths at 95 percent, mean 202 per row}} \\
\multicolumn{6}{l}{\textsc{Fixed lengths}} \\
\quad Per-POS lengths & 1,665 & 0.362\,/\,0.300 & 0.462\,/\,0.436 & 0.435\,/\,0.521 & 0.570\,/\,0.599 \\
\quad Shared length, 202 & 1,667 & 0.363\,/\,0.302 & 0.462\,/\,0.435 & 0.451\,/\,0.534 & 0.571\,/\,0.602 \\
\bottomrule
\end{tabular*}
\end{center}
\end{table}

Spending the budget by part of speech instead of by frequency does worse, except on rare words (Tables~\ref{tab:alloc} and \ref{tab:matched}). One length per POS class, 148 for verbs, 169 for nouns, 166 for proper nouns, 163 for adjectives, 108 for adverbs and 82 for other words, a mean of 163 per row, is no better than one shared length of 163 dimensions at the same storage, about 1,347 MB or 55 percent of the full table (Table~\ref{tab:matched}): it matches the shared length on SimVerb-3500 and SimLex-999 nouns and lowers verb-to-noun MRR by 0.028 and 0.023. Verbs, nouns, proper nouns and adjectives hold 98 percent of the storage, and their lengths, 148 to 169, all lie near the shared 163, so the two tables differ little outside adverbs and other words, 2 percent of the storage. At a mean of 202 per row, 1,665 MB, the per-POS lengths again match a shared length of the same storage on both sets and lower verb-to-noun MRR by 0.016 and 0.013. Allocation by POS class only, in which the greedy rule weighs a class by its tokens per row, 333 for proper nouns against 1,760 for nouns and 7,655 for verbs, also beats one shared length but falls short of the allocation by frequency: at a mean of 25 dimensions it is 0.079 and 0.063 above the POS bases at a uniform 25 per row on SimVerb-3500 (Table~\ref{tab:alloc}) but 0.055 and 0.044 below the allocation by frequency on SimLex-999 nouns, and at a mean of 100 it is 0.034 and 0.029 below the full table on MEN with all pairs, because it starves proper nouns. On RW nouns it scores 0.009 to 0.041 above the allocation by frequency at every budget, because it gives the rare words of a class as many dimensions as the frequent ones. Both ablations give each class one length and differ only in how it is set. Set from the class's own variance, the lengths match one shared length; set from its tokens per row, they beat it, and at 1,240 MB the allocation by POS class only scores 0.371 and 0.305 on SimVerb-3500 against 0.354 and 0.293 for the per-POS lengths at 1,348 MB. What helps is how often the words occur, not their part of speech, and the allocation by frequency uses that directly.

The allocation by frequency keeps the full table's similarity on SimVerb-3500 and SimLex-999 nouns from 829 MB on (Table~\ref{tab:alloc}), which shows that these benchmarks score mostly frequent words; at 1,347 MB, the storage of the per-POS lengths (Table~\ref{tab:matched}), SimVerb-3500 is 0.371 and 0.304 and SimLex-999 nouns 0.470 and 0.439. Nouns need more dimensions than verbs clearly only among words seen at least 10,000 times, and by a few directions between 1,000 and 10,000 (Section~\ref{sec:intrinsic}); at every budget of Table~\ref{tab:alloc} the allocation keeps the rows seen at least 1,000 times, 3 percent of the table, at 300 dimensions (299 for one block in window 5 at a mean of 25), as a rule that counts each word as often as it occurs is built to do; the rows it cuts are seen fewer than 1,000 times, and among words seen 100 to 1,000 times the POS classes need about the same number of dimensions. At a mean of 5 dimensions per row, below the budgets of Table~\ref{tab:alloc}, the frequent rows are cut too, and the POS split then matters in both directions: against the allocation by frequency only it raises SimVerb-3500 by 0.016 and 0.014 and lowers MEN with all pairs by 0.014 and 0.017; at a mean of 10 the two differ by at most 0.003 on these sets. The allocation also keeps the verb-to-noun retrieval that the per-POS lengths lose: MRR is 0.496 and 0.556 against 0.365 and 0.469 for the per-POS lengths and 0.479 and 0.569 for the full table, so the per-POS lengths lose retrieval by cutting frequent rows, not through the basis, which is the same. Rare-word neighbors are mostly a matter of the basis: rare-word neighbor retention (R1) is 0.464 and 0.488 with the POS bases and 0.502 and 0.529 with the per-POS lengths, against 0.640 and 0.690 with the band bases at the same storage.

The basis matters where rows are cut. At one fixed length every row is cut, and the POS bases describe verbs, adjectives and nouns better than the band bases (Table~\ref{tab:bases}: 0.339 against 0.265 on SimVerb-3500 at $k = 100$, $w{=}1$). This is not at odds with Section~\ref{sec:crosspos}, where at one fixed length the POS bases lower verb-to-noun MRR to 0.245 against 0.401 for the global basis: there every row is truncated in its class's subspace, which removes relations between classes. Under the allocation only rare words are cut, and the two bases divide what they keep of them: the POS bases, weighted toward frequent words (Section~\ref{sec:weighting}), keep verb-to-noun retrieval, with an MRR of 0.496 and 0.556 against 0.452 and 0.550 for the band bases at 1,347 MB (Table~\ref{tab:matched}), whereas the band bases, fitted to the rare words themselves, keep more of the rare words, up to 0.015 on RW nouns (Table~\ref{tab:alloc}) and R1 0.640 and 0.690 against 0.464 and 0.488. We therefore allocate with the POS bases and switch to the band bases only where the neighbors of rare words matter more than verb-to-noun retrieval.

In short, part of speech decides the directions a word keeps, which matters where rows are cut, and frequency how many: from a third of the table's storage on, the POS bases with a budget spent by frequency keep the full table's similarity on SimVerb-3500 and SimLex-999 nouns and its verb-to-noun retrieval within 0.02, no part of speech needs a length of its own, and against one basis per frequency band the POS bases keep up to 0.04 more of that retrieval and lose only on the rare words.

\section{Conclusion}\label{sec:conclusion}

During training, POS tags help verbs, raising SimVerb-3500 by 0.016 to 0.033 at every dimension and window, and the vectors must stay separate: averaging a word's vectors back into one falls below the original table on MEN with all pairs. Verbs also gain more from the adverbs around them than from as many random words. Nouns gain much less from POS tags, and adjectives give them only a small gain over random words.

In the trained space, the POS bases, one PCA basis per POS class, serve verbs and adjectives better: at one shared length of 25 to 150 dimensions they score higher than the global basis or the band bases on SimVerb-3500 and SimLex-999 adjectives, with or without count weighting. For nouns, however, most of the advantage over the global basis at 100 dimensions comes from count weighting. With both count-weighted, the POS bases at 100 dimensions also keep more verb and noun neighbors in the query's WordNet semantic category than the global basis.

Among words seen at least 10,000 times, nouns need 12 to 22 more directions than verbs at 90 percent of the variance, and 3 to 7 more between 1,000 and 10,000 times; verbs need more than other words in every band and window, and the same order holds at 95 and 99 percent. For words seen 100 to 1,000 times the POS classes differ little.

A budget of dimensions is best spent by frequency. With the POS bases and a greedy allocation by training count, from a third of the table's storage on only words seen fewer than 100 times lose dimensions, SimVerb-3500 and SimLex-999 nouns stay at the full table's level and verb-to-noun retrieval stays within 0.02, while similarity on rare nouns falls by up to 0.04 and at 829 and 1,347 MB rare words keep 30 to 49 percent of their nearest neighbors. Nouns need more dimensions than verbs only among frequent words (Section~\ref{sec:geometry}), and those keep at least 299 of their 300 dimensions under the budgets we test, so at these budgets no part of speech needs a length of its own; where frequent words are cut too, at a mean of 5 dimensions per row, the POS split gains on SimVerb-3500 and loses on MEN (Section~\ref{sec:method-results}). The basis matters only for the rows that are cut: at one shared length the POS bases score higher than the global basis and the band bases (Section~\ref{sec:subspaces}), and under the allocation they keep up to 0.04 more of the verb-to-noun retrieval than one basis per frequency band and lose only on the rare words. When a table must lose dimensions, we therefore recommend the POS bases with the budget spent by frequency, and one basis per frequency band only where the neighbors of rare words matter more than verb-to-noun retrieval. Part of speech decides what a word's vector should be, through the contexts it is trained on and, where rows are cut to one length, the subspace that describes it; frequency decides how much of it to keep.

\FloatBarrier
\printbibliography
\end{document}